\def\year{2021}\relax
\documentclass[letterpaper]{article}
\usepackage{aaai20}
\usepackage{times}
\usepackage{helvet}
\usepackage{courier}
\usepackage[hyphens]{url}
\usepackage{graphicx}
\usepackage{xcolor}

\usepackage{graphicx}
\usepackage{amsmath}
\usepackage{listings}
\title{Goal Reasoning by Selecting Subgoals with Deep Q-Learning}
\author{Carlos Núñez-Molina,\textsuperscript{\rm 1} Vladislav Nikolov,\textsuperscript{\rm 2}, Ignacio Vellido,\textsuperscript{\rm 3} Juan Fernández-Olivares,\textsuperscript{\rm 4} \\
Departamento de Ciencias de la Computación e IA, Universidad de Granada\\
Calle Periodista Daniel Saucedo Aranda, Granada 18014\\
\textsuperscript{\rm 1}ccaarlos@correo.ugr.es, \textsuperscript{\rm 2}vladis890@correo.ugr.es, \textsuperscript{\rm 3}ignaciove@correo.ugr.es, \textsuperscript{\rm 4}faro@decsai.ugr.es, \\
} 
\begin{document}
\maketitle
\begin{abstract}
In this work we propose a goal reasoning method which learns to select subgoals with Deep Q-Learning in order to decrease the load of a planner when faced with scenarios with tight time restrictions, such as online execution systems. We have designed a CNN-based goal selection module and trained it on a standard video game environment, testing it on different games (planning domains) and levels (planning problems) to measure its generalization abilities. When comparing its performance with a satisfying planner, the results obtained show both approaches are able to find plans of good quality, but our method greatly decreases planning time. We conclude our approach can be successfully applied to different types of domains (games), and shows good generalization properties when evaluated on new levels (problems) of the same game (domain). 
\end{abstract}

\section{Introduction}
Automated Planning has traditionally been one of the most widely used techniques in AI and has been successfully applied in real-world applications  \cite{castillo2008samap,fdez2019personalized}. However, in order to integrate it into online execution systems, i.e., systems used in real-time scenarios which interleave planning and acting, there exist several issues which must be addressed. Firstly, planning is often too slow for real-time scenarios. In most real-world problems the search space is enormous so, despite the use of heuristics, finding a suitable plan usually takes very long. Secondly, since most real-world environments are highly dynamic, it is very likely that the environment has changed before a long plan has finished being executed.

Despite great advances in the integration of planning and acting into online architectures \cite{patra2019acting,ingrand2017deliberation,guzman2012pelea,Niemueller_Hofmann_Lakemeyer_2019}, the above features still hinder the generalized adoption of automated planning in such scenarios. Because of that, many recent works which apply AI to guide agents behaviour in real-time scenarios, like video games, choose to rely on Machine Learning alone and do not integrate planning into their agent architecture. This can be clearly seen in \cite{vinyals2019grandmaster}. In this impactful work, an agent is trained to play \emph{Starcraft}, a highly competitive real-time strategy (RTS) game. This seems like a perfect problem for planning: players need to establish a long-term, goal-oriented strategy in order to achieve victory and all the dynamics of the game are known, so they can be represented into a planning domain. However, Vinyals et al. choose to integrate Deep Learning \cite{lecun2015deep} with Reinforcement Learning \cite{sutton2018reinforcement} to model the behaviour of the agent.

Architectures which rely on Machine Learning (ML) and Reinforcement Learning (RL) present some advantages over planning: they usually require very little prior knowledge about the domain (they do not need a planning domain) and, once trained, they act quickly, since they do not perform any type of planning. Nevertheless, they also have some drawbacks. Firstly, they are very sample inefficient. They require a lot of data in order to learn, in the order of hundreds of thousands or even millions samples \cite{torrado2018deep}. Secondly, they usually present bad generalization properties, i.e., have difficulties in applying what they have learnt not only to new domains but also to new problems of the same domain \cite{zhang2018study}.

Since both Automated Planning and Reinforcement Learning have their own pros and cons, it seems natural to try to combine them as part of the same agent architecture, which ideally would possess the best of both worlds. For that purpose, we have resorted to \emph{Goal Reasoning} \cite{aha2015goal}, a design philosophy for agents in which its entire behaviour revolves around goals. They learn to formulate goals, select goals, achieve the selected goals and select new goals when \emph{discrepancies} are detected.

The main contribution of this paper is the proposal of a RL-based Goal Selection Module and its integration into a planning and acting architecture to control the behaviour of an agent in a real-time environment. We have trained and tested our approach on the GVGAI video game framework \cite{perez20152014}. GVGAI is a framework intended to evaluate the behaviour of reactive and deliberative agents in several video games. Its ultimate goal is to help advance the state of the art in General Artificial Intelligence. 

The Goal Selection Module here presented is based on a Convolutional Neural Network (CNN) \cite{krizhevsky2012imagenet} which has been trained with the RL algorithm known as Deep Q-Learning \cite{mnih2013playing}. The training experience has been extracted from the execution of thousands of episodes of a planning agent that randomly selects subgoals in the GVGAI environment, on both, different domains and  different problems for each domain. Training problems are also different from the ones used for testing, which allows us to evaluate the generalization ability of the module with respect to both domains and problems.

The CNN receives as input an image-like encoding of the current state of the game $s$ and an eligible subgoal $g$ and returns the predicted length of the plan which starts at $s$, achieves $g$ and then achieves the final goal (wins the game). The Goal Selection Module selects the subgoal $g^*$ whose associated plan has the minimum predicted length. After selecting $g^*$, the Planner Module finds a valid plan from $s$ to $g^*$, which will then be executed by the agent in GVGAI. 

We have conducted an experimentation to evaluate the total planning time taken by our approach, with respect to the planning time taken to produce the first solution to every  original problem with a satisfying planner \footnote{We have used FF as the baseline planner.}. Our experimentation also shows a comparison of the quality of plans produced by both approaches. The results obtained show both approaches are able to find plans of good quality, but our method greatly decreases planning time when applied to complex problems. Moreover, we have observed in our experiments that using our approach  planning time remains almost constant for complex problems where our baseline satisfying planner fails to find a solution in reasonable time. We think that this is an argument that can favour the adoption of planning integrated with goal selection in scenarios with tight time restrictions.

Addressing Goal Selection with Deep Q-Learning and a CNN has two main advantages. Firstly, as the results of our experiments show, the Goal Selection Module learns to generalize. The use of a CNN allows it to apply what has learnt on the training levels to new levels it has never seen before. Secondly, thanks to the use of Deep Q-Learning, the Goal Selection Module learns to select goals \emph{thinking in the long term}, i.e, taking into account the subgoals it will have to achieve afterwards to beat the game.

The structure of this work is the following. We first explain the GVGAI framework and the Deep Q-Learning algorithm. We then present an overview of the architecture and show how the Goal Selection Module learns. After that, we present the results of our empirical study. We then compare our approach with related work. We finish by presenting our conclusions and future work.

\section{Background}

\subsection{GVGAI}
To test our planning and acting architecture we have used the General Video Game AI (GVGAI) Framework \cite{perez20152014}. This framework provides a game environment with a large quantity of tile-based games which are also very different in kind. For example, it comprises purely reactive games, such as \emph{Space Invaders}, and also games which require long-term planning in order to be solved successfully, such as \emph{Sokoban}.
We have chosen to use deterministic versions of three GVGAI games (known as \textit{Boulder Dash, IceAndFire, and Catapults} detailed in the experiments section). We use these games to extract the experience of episodes of planning and acting our Goal Selection Module is trained on. All the games require both deliberation and long term thinking to be solved. All of them share that it is necessary to reach an exit portal after accomplishing some subgoals which involve gathering objects on given cells.

\begin{figure}[h]
	\centering
	\includegraphics[width=\linewidth]{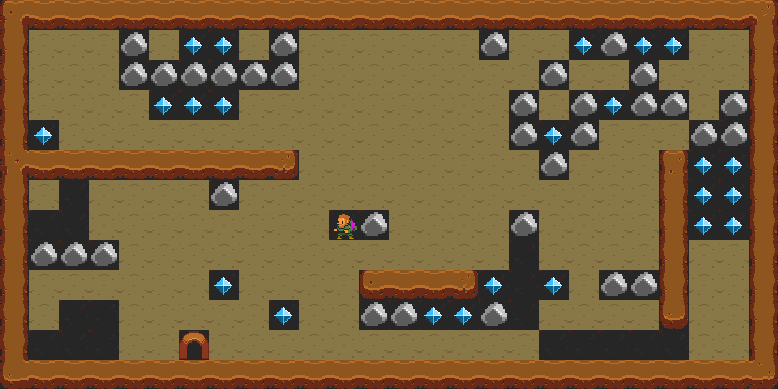}
	\caption{A level of the BoulderDash game.}
	\label{fig:boulder_dash}
\end{figure}

As an example, Figure \ref{fig:boulder_dash} shows the configuration of a level in the game \textit{Boulder Dash}. In our version of Boulder Dash, the player must collect nine gems and then go to the exit, while minimizing the number of actions used. In order to do that, it must traverse the level (one tile at a time) while overcoming the obstacles: the player cannot pass through walls and boulders must be broken with its pickaxe before passing through. Also, the player must select which gems to collect, since there are more than nine gems available.  All of this makes it really hard to find the shortest plan, even a first solution plan for a satisfying planner, as shown in the experiments. 

One very important reason we have chosen GVGAI is because it makes available a mechanism for easily creating and integrating new games and levels. This way, we can create as many new levels for a given game as we want, which allows us to test the generalization abilities of a planning and acting architecture when its Goal Selection Module has already been trained. The Video Game Description Language, VGDL \cite{perez20152014}, is the method used to define the dynamics and interactions of all the objects in each of the games. Every level in the game is defined by a level description file, which contains the layout of the level and the initial positions of the objects. Listing 1 shows the associated level description file of the game level shown on Figure \ref{fig:boulder_dash}. Each type of object has a character associated: \emph{w} for walls, \emph{o} for boulders, \emph{x} for gems, \emph{A} for the player, \emph{e} for the exit, \emph{.} for tiles and \emph{-} for empty tiles, which are the same as normal tiles.

\newpage

\begin{center}
  \lstset{
    caption=The level description file of the level shown on Figure \ref{fig:boulder_dash}.,
    captionpos=b
  }
  \begin{lstlisting}
	wwwwwwwwwwwwwwwwwwwwwwwwww
	w...o.xx.o......o..xoxx..w
	w...oooooo........o..o...w
	w....xxx.........o.oxoo.ow
	wx...............oxo...oow
	wwwwwwwwww........o...wxxw
	w.-....o..............wxxw
	w--........Ao....o....wxxw
	wooo.............-....w..w
	w......x....wwwwx-x.oow..w
	w.--.....x..ooxxo-....w..w
	w---..e...........-----..w
	wwwwwwwwwwwwwwwwwwwwwwwwww
  \end{lstlisting}
\end{center}

\subsection{Deep Q-Learning}
Q-Learning \cite{watkins1989learning} is one of the most widely used techniques in Reinforcement Learning, RL, \cite{sutton2018reinforcement}. As every RL technique, it learns a policy $\pi$ that, in every state $s$, selects the best action $a$ in the set of available actions $A$ in order to maximize the expected cumulative reward $R$, i.e., the expected sum of all the (discounted) rewards $r$ obtained by choosing actions according to the same policy $\pi$ from the current state $s$ until the end of the episode. According to the \emph{Reward Hypothesis}, all goals can be described as the maximization of $R$. This means that, no matter the goal an agent is pursuing, its behaviour can be modeled and learnt (more or less successfully) using a RL technique, such as Q-Learning.

Q-Learning associates a value to each $(s,a)$ pair, known as the Q-value, $Q(s,a)$. This value represents the expected cumulative reward $R$ associated with executing action $a$ in state $s$, i.e., how good $a$ is when applied in $s$. This way, the policy $\pi$ learnt with Q-Learning corresponds to, given a state $s$, selecting the action $a^*$ in $A$ with the maximum Q-value associated.

One of the main problems Q-Learning has is that it needs to learn the associated Q-value for each of the $(s,a)$ pairs, known as the \emph{Q-table}. If the action or state space are too big, the Q-table grows and the learning problem becomes intractable. Deep Q-Learning \cite{mnih2013playing} solves this problem. Instead of learning the Q-table, it uses a Deep Neural Network (DNN) to learn the Q-values. Thanks to the use of a DNN, it is able to generalize and correctly predict the Q-values for new $(s,a)$ pairs never seen before by the network. In our work, we select the best subgoal from a set of possible subgoals. The set of possible subgoals depends on the current state $s$. Since the state space is enormous, the size of the set of possible subgoals across all different states is also really big. For this reason, we use Deep Q-Learning in pursuit of the good generalization abilities shown by \cite{mnih2013playing}.

\section{The Planning and Acting Architecture}

\begin{figure}[h]
	\centering
	\includegraphics[width=\linewidth]{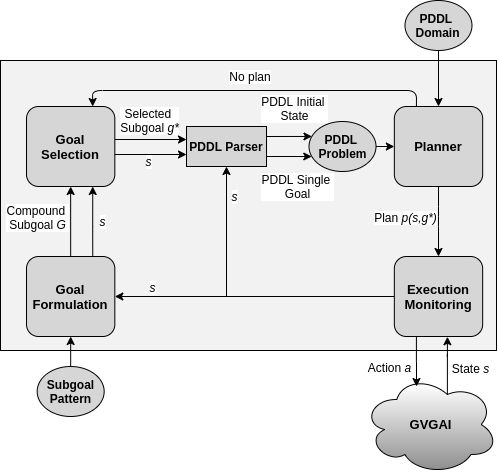}
	\caption{An overview of the planning and acting architecture.}
	\label{fig:architecture}
\end{figure}

An overview of the planning and acting architecture can be seen on Figure \ref{fig:architecture}. The \textbf{Execution Monitoring} Module communicates with the GVGAI environment, receiving the current state $s$ of the game. It also supervises the state of the current plan. If it is not empty, it returns the next action $a$. If it is empty, the architecture needs to find a new plan. 

The \textbf{Goal Formulation} Module receives $s$ and generates the compound subgoal $G$, which is a list of single subgoals $\{g_1, g_2, ..., g_n\}$. Since all GVGAI games are tile-based, we have associated each subgoal with getting to its correspondent tile (cell), which permits to handle subgoals for any of the games represented in this work. The \textbf{Subgoal Pattern} contains the prior information about each game domain needed to automatically generate $G$ given $s$.  It is encoded as a list of  object classes that correspond to subgoals. This allows us to easily adapt the \textit{Goal Formulation} to a new GVGAI game since we only need to provide the \textit{Subgoal Pattern} with a list of object classes corresponding to subgoals in this new domain.

In every game each subgoal $g \in G$ corresponds to getting to a level tile that contains an object of the classes defined in the \textit{Subgoal Pattern}  or, if the player has already achieved all the necessary subgoals, the final goal $g_f$ (get to the exit) is directly attainable and $G=\{g_f\}$. The \textbf{Goal Selection} Module receives $G$ and selects the best subgoal $g^* \in G$ given $s$ (the mechanism is explained in the next section).

The \textbf{PDDL Parser} encodes $g^*$ as a PDDL Single Goal, i.e., (\emph{goto tile13}), and $s$ as a PDDL Initial State, which together constitute the PDDL Problem. The \textbf{Planner} Module receives the PDDL Problem along with the PDDL Domain, provided by a human expert, and generates a plan $p(s,g^*)$ which achieves $g^*$ starting from $s$. Finally, the \textit{ Execution Monitoring Module} receives $p(s,g^*)$ and the cycle completes. It is worth noting that the list of subgoals received by the \textit{Goal Selection Module} might contain either unreachable or dead-end subgoals (the player dies). In the first case, the planner cannot find a plan and notifies that situation to the \textit{Goal Selection Module}, that selects the next best subgoal. In the second case, the agent fails to solve the problem. As explained in the following, the Deep Q-Learning learns to not select these types of subgoals.

\section{Goal Selection Learning}
In order to select the best subgoal $g^* \in G$ for a given $s$, the Goal Selection Module iterates over every $g \in G$ and predicts the length of its associated plan. It then selects as $g^*$ the subgoal whose associated plan has been predicted the minimum length. The Module uses a Convolutional Neural Network (CNN) \cite{krizhevsky2012imagenet} that receives $s$ and a $g \in G$, both encoded as a \emph{one-hot matrix}, and outputs the predicted plan length. Each position of this one-hot matrix corresponds to a tile of the level of a game, and encodes the objects within that tile as a \emph{one-hot vector}, i.e., a vector where each position is associated with a different type of object and which contains \emph{1} if that object is in that tile and \emph{0} otherwise. The subgoal $g$ is also encoded in the one-hot vector of its associated tile. 

Our approach for Goal Selection uses a Deep Q-Learning based model (which we call \textbf{DQP Model}, an acronym for \textit{Deep-Q Planning}) that predicts as $l_{P(s,g)}$ the length of the plan $P(s,g)$ that achieves $g$ \emph{and}, after reaching it, achieves the final goal $g_f$ (after obtaining all the required subgoals in an optimal way). This way, the DQP Model predicts the length of the entire plan, not only the first section of it, that we note as $p(s,g)$, which corresponds to a plan that achieves $g$ starting from $s$.

Since only the length of the first section of the  plan $p(s,g)$ is known, this model cannot be trained in a supervised fashion, since the length of the plan that achieves subgoals in an optimal way is unknown. To train this model, we have chosen to apply the methodology followed by Deep Q-Learning \cite{mnih2013playing}. To do so, we establish a correspondence between our problem and Reinforcement Learning (RL). Actions $a$ in RL correspond in our work to achieving a subgoal $g$, the reward $r$ obtained by executing $a$ at $s$ corresponds to the length of the plan $p_{s,g}$ that starts at $s$ and achieves a subgoal $g$, the expected cumulative reward $R$ associated with $(s,a)$ corresponds to the length $l_{P(s,g)}$ of the entire plan $P_{s,g}$, and maximizing $R$ corresponds to minimizing $l_{P(s,g)}$. Table \ref{table:RL_comparison} shows this correspondence. Moreover, when $g$ corresponds to an \textit{unreachable} or \textit{dead-end} goal (explained above), $r$ = 100, while $r$=-100 for $g$ being the final goal. This way we are representing a penalty (a really long plan length) to avoid unreachable or dead-end goals, and a big reward (a plan of \textit{negative } length) for the final goal, thus allowing the agent to learn to reject \textit{bad} goals and to select the final goal as soon as it is attainable.

\begin{table}[h]
\label{table:RL_comparison}
\centering
\begin{tabular}{|c|c|}
\hline
\textbf{RL}                        & \textbf{Our Work}         \\ \hline
Action $a$         				   & Subgoal $g$ 				   \\ \hline
Reward $r$          				   & $l_{p(s,g)}$ \\ \hline
Cumulative Reward $R$          	   & $l_{P(s,g)}$         \\ \hline
Maximize $R$ 	  			   		   & Minimize $l_{P(s,g)}$       \\ \hline
\end{tabular}
\caption{Correspondence between RL and our problem.}
\end{table}

The CNN of the DQP Model predicts $l_{P(s,g)}$, which in Deep Q-Learning corresponds to the Q-value $Q(s,a)$. Since its correct value, the Q-target $Q^*(s,a)$, is unknown, it is estimated using other predicted Q-values $Q(s',a')$ in a technique known as \emph{bootstrapping}. This is the method used to learn the Q-values. The network is trained by minimizing the squared difference between $Q(s,a)$ and $Q^*(s,a)$. This loss $L$ formula is called the Bellman Equation and is shown below:

\begin{equation*} 
L = (Q(s,a) - Q^*(s,a))^2 = 
\end{equation*} 
\begin{equation} 
(Q(s,a) - (r + \gamma \max_{a' \in A'} Q(s',a')))^2
\end{equation}

\noindent where $s'$ is the next state (after applying $a$ in $s$), $A'$ is the set of applicable actions in $s'$ and $\gamma=1$ is the \emph{discount factor}, so actually we don't discount future rewards (plan lengths).

The CNN architecture used for the DQP Model is composed of 8 convolutional layers and 2 inner fully connected (fc) layers, without considering the output layer. The first two convolutional layers contain 32 filters each one, the next three use 64 filters each, and the last three layers use 128 filters each one. Then, the first fc layer contains 128 units and the next fc layer 32 units. We normalized the dataset before using it to train the CNN. Also, in order to make learning more stable, an auxiliary, independent CNN is used to estimate the Q-targets, in a technique known as \emph{Fixed Q-targets} \cite{mnih2015human}. 

The  DQP model use \emph{offline learning}, i.e., is trained on static datasets. These datasets are populated by performing \emph{random exploration} on the training levels of the corresponding game. Each time the Goal Selection Module must select a new subgoal $g^*$ for the current state $s$, it selects it randomly. Then, when the architecture has found $p(s,g^*)$ and executed it arriving at state $s'$, a new sample is added to the datasets. The datasets of the DQP Model are filled with samples of the form $(s,g^*,r,s')$.

\section{Experiments and Analysis of Results}

We have conducted an experimentation with a two-fold goal in mind: (1) to test the  generalization abilities of our DQP model, by training and testing it on different levels and domains, (2) to compare the total  time (planning time + goal selection time) taken by our approach, with respect to the planning time needed by a classical planner using different optimization options.

We have trained and tested our approach on three different GVGAI games: BoulderDash, IceAndFire and Catapults. The (final) goal of every game is getting to the exit after meeting certain requirements, i.e., achieving several subgoals, while minimizing the number of actions used. In our deterministic version of BoulderDash, the agent must traverse the level, collect at least nine gems and then get to the exit. In this game there are two types of obstacles: boulders, which must be broken with a pickaxe before passing through, and walls, which are impassable. Subgoals in BoulderDash correspond to items of the class \emph{gem}. This information is encoded in the \textit{Subgoal Pattern Module} for the architecture to be able to correctly formulate subgoals. In IceAndFire, the agent must traverse the level, collect the ten coins present at the map and get to the exit. In this game there are impassable obstacles (walls and spikes) but, unlike BoulderDash, there are also tiles with ice and fire which can only be traversed after obtaining ice boots and fire boots, respectively. Thus, subgoals correspond to items of the class \emph{coin}, \emph{fire-boots} and \emph{ice-boots}, which must be pursued in the right order so as to correctly avoid the obstacles. In Catapults, the agent must use the catapults in order to get to the exit safely. There are four types of catapults (\emph{up}, \emph{right}, \emph{left} and \emph{down}), which correspond to the subgoals in this game. When the agent steps on a catapult, it is launched towards the corresponding direction and keeps flying until it hits a wall or another catapult, in which case this process repeats recursively. If the tile where the agent ends after this flight contains water, the agent dies and loses the level automatically, therefore the model has to learn to avoid these subgoals. Another way of losing this game is getting to a dead-end state, i.e., a state from which no subgoal (catapult) or final goal (exit) is achievable. This is why Catapults is the hardest of the three games: the agent must carefully select the correct catapults and in the right order so as to get to the exit without dying.

For each game, we have represented a PDDL planning domain and we have collected datasets to train our architecture on. To do this, the agent, making use of the \textit{Planning and Action Architecture}, performed random exploration, i.e., the \textit{Goal Selection Module} selected subgoals at random and sent them to the planner, on the training levels of each game. For each level, we saved all the samples collected by the agent up to 500 unique (non-repeating) samples per level or all the unique samples obtained after 1000 iterations, since there are levels which don't contain so many unique samples. We have used 100 training levels for BoulderDash and IceAndFire and 200 levels for Catapults (we are using VGDL along with a GUI-based tool to easily create new levels), since we have extracted fewer samples for each level of this game. In total, this accounts for 50000 training samples in BoulderDash, 42950 in IceAndFire and 60018 in Catapults. 

These datasets were not only used to train the Planning and Acting architecture but also to select and validate different CNN architectures and hyperparameters for the Goal Selection Module. This was made by training the candidate CNN architectures on a subset of the training dataset and evaluating their performance on levels not used for training. This way, we selected the best CNN architecture which is the same one for the three games, except for the fact that we apply Batch Normalization after every convolutional layer for BoulderDash\footnote{All the resources, software, planning domains, and  levels used including the test levels are available online in a public repository that will be provided should the paper be accepted. }.

Once we obtained the best CNN architecture, we trained one DQP model on the entire training dataset for each game. We used 20000 training iterations for BoulderDash and IceAndFire and 25000 for Catapults. Each trained model was evaluated on the test levels. These test levels were different from the ones used for training in order to measure the generalization ability or our approach when applied to levels never seen before. The performance of our architecture was measured according to the length (number of actions) of the plans obtained and the time needed to obtain them (goal selection and planning times). In Catapults, since the agent can die, we also measure the \emph{success rate}, i.e., how often the agent can complete each level (without dying).

We have chosen the Fast-Forward (FF) Planning System \cite{hoffmann2001ff} for our \textit{Planner Module} since the version of PDDL its parser uses is expressive enough to represent domains such as those of video games. We have selected the Best-First-Search (BFS) with  $g=1$ and $h=5$ as the search strategy for FF when planning for a given subgoal. This way, FF finds a valid plan which achieves the subgoal, trying to minimize its number of actions although it is not guaranteed to obtain the shortest possible plan.

In order to compare the performance of our Planning and Acting architecture with respect to classical planning, we tried to solve the same test levels using FF but, this time, without employing our architecture. This means we executed FF on the PDDL problem associated with each test level, solving it completely with no goal selection whatsoever, as in classical planning. We tried to obtain the optimal (shortest) plan for each level using the BFS strategy with $g=1$ and $h=1$ but, since many levels were too complex for FF to solve optimally, we also executed FF with \emph{soft} optimization options (BFS with options $g=1$ and $h=5$, as used when performing goal selection) and with no optimization options at all, making use in this case of the Enforced-Hill-Climbing (EHC) search strategy.

Lastly, in order to assess the quality of the goal selection performed by our approach, we compared it with a model which selects subgoals completely at random, which we call Random Model. This baseline model corresponds to using the Planning and Acting architecture but, instead of employing the Goal Selection Module to select subgoals, it selects them at random. This way, the Random Model represents the worst possible way of selecting subgoals.

The test levels used to compare the performance of the different techniques were comprised of the five levels provided by default in GVGAI for each game and also 4 new levels we created. These additional test levels (which will be referred to as \emph{hard} levels) were purposely created so that they were more complex and harder to solve by FF, but of the same size, i.e., number of tiles, as the other test levels (which will be referred to as \emph{easy} levels). For instance, in BoulderDash we discovered that FF had trouble solving levels which contained a lot of boulders.

Tables 2, 3 and 4 show the performance obtained by the different approaches on both the easy and hard levels for each game. For the Planning and Acting architecture and the Random Model, we repeated each execution 15 times and averaged the results. For the FF planner, we repeated each execution 5 times for every search strategy and averaged the planning times. We allowed FF to spend a maximum of 1 hour of planning time for each level. If after this time FF had not found a plan yet, we considered the corresponding level as too complex for FF to solve.

\begin{table}[h]
	\label{table:BoulderDashResults}
	\centering
	\resizebox{.48\textwidth}{!}{
		\begin{tabular}{|l|l|l|l|l|l||l|l|l|l|}
			\hline
			\multicolumn{10}{|c|}{\textbf{Number of Actions in BoulderDash}}                             \\
			\hline
			& \multicolumn{5}{c||}{Easy Levels} & \multicolumn{4}{c|}{Hard Levels} \\
			\hline
			& 0    & 1    & 2    & 3    & 4   & 5      & 6      & 7     & 8     \\
			\hline
			DQP     & 99   & 53   & 62   & 93   & 70  & 108    & \textbf{142}    & \textbf{98}    & 121   \\
			Random  & 207  & 188  & 144  & 177  & 190 & 214    & 302    & 239   & 262   \\
			Optimal & -    & 31   & 42   & -    & 38  & -      & -      & -     & -     \\
			BFS     & 80   & 51   & 42   & 74   & 41  & -      & -      & -     & 114   \\
			EHC     & -    & 46   & 41   & 83   & 53  & 106    & -      & -     & -     \\
			\hline
		\end{tabular}
	}
	
	\vspace{.3cm}
	
	\resizebox{.48\textwidth}{!}{
		\begin{tabular}{|l|l|l|l|l|l||l|l|l|l|}
			\hline
			\multicolumn{10}{|c|}{\textbf{Total Time (s) in BoulderDash}}                                     \\
			\hline
			& \multicolumn{5}{c||}{Easy Levels}      & \multicolumn{4}{c|}{Hard Levels} \\
			\hline
			& 0     & 1    & 2     & 3     & 4     & 5       & 6     & 7     & 8     \\
			\hline
			DQP     & 1.90  & 0.63 & 1.60  & 0.79  & 1.60  & \textbf{0.89}    & \textbf{1.82}  & \textbf{0.86}  & \textbf{1.85}  \\
			Random  & 0.47  & 0.43 & 0.32  & 0.39  & 0.45  & 0.55    & 0.61  & 0.53  & 0.52  \\
			Optimal & -     & 8.41 & 15.83 & -     & 407.6 & \textbf{-}       & \textbf{-}     & \textbf{-}     & \textbf{-}     \\
			BFS     & 231.0 & 0.75 & 0.04  & 667.6 & 0.06  & \textbf{-}       & \textbf{-}     & \textbf{-}     & \textbf{85.6}  \\
			EHC     & -     & 0.10 & 0.04  & 0.27  & 0.06  & \textbf{682.0}   & \textbf{-}     & \textbf{-}     & \textbf{-} \\
			\hline   
		\end{tabular}
	}
	
	\caption{Results obtained by each approach in BoulderDash. The symbol ``-" represents a timeout (FF could not find a plan in 1 hour).}
\end{table}

\begin{table}[h]
	\label{table:IceAndFireResults}
	\centering
	\resizebox{.48\textwidth}{!}{
		\begin{tabular}{|l|l|l|l|l|l||l|l|l|l|}
			\hline
			\multicolumn{10}{|c|}{\textbf{Number of Actions in IceAndFire}}  \\
			\hline 
			& \multicolumn{5}{c||}{Easy Levels} & \multicolumn{4}{c|}{Hard Levels} \\
			\hline 
			& 0    & 1    & 2    & 3    & 4   & 5      & 6      & 7     & 8    \\
			\hline 
			DQP     & \textbf{115}  & 109  & \textbf{110}  & \textbf{111}  & 167 & 111    & 181    & \textbf{111}   & 122   \\
			
			Random  & 140  & 109  & 117  & 135  & 182 & 117    & 181    & 114   & 143   \\
			
			Optimal & 84   & 83   & 97   & 89   & 126 & 78     & 128    & 73    & 79    \\ 
			
			BFS     & 84   & 83   & \textbf{109}  & \textbf{119}  & 126 & 82     & 152    & \textbf{115}   & 113   \\
			
			EHC     & \textbf{134}  & 97   & \textbf{113}  & \textbf{157}  & 130 & 98     & 160    & \textbf{131}   & 107  \\
			\hline 
		\end{tabular}
	}
	
	\vspace{.3cm}	
	
	\resizebox{.48\textwidth}{!}{
		\begin{tabular}{|l|l|l|l|l|l||l|l|l|l|}
			\hline
			\multicolumn{10}{|c|}{\textbf{Total Time(s) in IceAndFire}}  \\
			\hline 
			& \multicolumn{5}{c||}{Easy Levels}  & \multicolumn{4}{c|}{Hard Levels} \\
			\hline
			& 0    & 1    & 2    & 3    & 4    & 5     & 6      & 7      & 8    \\
			\hline
			DQP     & 1.63 & 1.30 & 1.29 & 0.48 & 1.32 & 1.46  &\textbf{0.58}   & \textbf{0.60}   &\textbf{1.39}  \\
			Random  & 0.27 & 0.30 & 0.18 & 0.19 & 0.17 & 0.20  & 0.32   & 0.26   & 0.19  \\
			Optimal & 0.43 & 0.79 & 0.72 & 1.33 & 0.72 & 0.74  & \textbf{11.97}  & \textbf{11.07}  &\textbf{9.23}  \\
			BFS     & 0.01 & 0.02 & 0.01 & 0.03 & 0.02 & 0.15  & \textbf{7.91}   & \textbf{5.32}   & \textbf{5.98}  \\
			EHC     & 0.01 & 0.01 & 0.01 & 0.02 & 0.01 & 0.60  & 1.21   & 0.72   & 0.66 \\
			\hline
		\end{tabular}
	}
	\caption{Results obtained by each approach in IceAndFire.}
\end{table}

\begin{table}[h]
	\label{table:CatapultsResults}
	\resizebox{.48\textwidth}{!}{
		\begin{tabular}{|l|l|l|l|l|l||l|l|l|l|}
			\hline
			\multicolumn{10}{|c|}{\textbf{Success Rate (\%) in Catapults}}                                 \\
			\hline
			& \multicolumn{5}{c||}{Easy Levels}     & \multicolumn{4}{c|}{Hard Levels} \\
			\hline
			& 0    & 1     & 2     & 3     & 4    & 5      & 6     & 7      & 8     \\
			\hline
			DQP    & 0.0  & \textbf{26.66} & \textbf{33.33} & \textbf{26.66} & 0.0  & \textbf{20.0}   & 0.0   & \textbf{33.33}  & \textbf{6.66}  \\
			Random & 20.0 & \textbf{6.66}  & \textbf{13.33} & \textbf{20.0}  & 6.66 & \textbf{0.0}    & 0.0   & \textbf{0.0}    & \textbf{0.00} \\
			\hline
		\end{tabular}
	}
	
	\vspace{.3cm}	
	
	\resizebox{.48\textwidth}{!}{
		\begin{tabular}{|l|l|l|l|l|l||l|l|l|l|}
			\hline
			\multicolumn{10}{|c|}{\textbf{Number of Actions in Catapults}}  \\
			\hline
			& \multicolumn{5}{c||}{Easy Levels} & \multicolumn{4}{c|}{Hard Levels} \\
			\hline
			& 0    & 1    & 2    & 3    & 4   & 5       & 6    & 7      & 8     \\
			\hline
			DQP     & -    & \textbf{21}   & \textbf{15}   & \textbf{27}   & -   & 847     & -    & 147    & 92    \\
			Random  & 25   & 21   & 15   & 27   & 22  & -       & -    & -      & -     \\
			Optimal & 25   &\textbf{21}   & 13   & \textbf{27}   & 22  & -       & -    & -      & -     \\
			BFS     & 27   & \textbf{21}   & \textbf{17}   & \textbf{29}   & 22  & -       & -    & -      & 42    \\
			EHC     & 27   & \textbf{23}   & 13   & \textbf{29}   & 22  & 277     & -    & -      & -   \\
			\hline 
		\end{tabular}
	}
	
	\vspace{.3cm}
	
	\resizebox{.48\textwidth}{!}{
		\begin{tabular}{|l|l|l|l|l|l||l|l|l|l|}
			\hline
			\multicolumn{10}{|c|}{\textbf{Total Time(s) in Catapults}}  \\
			\hline 
			& \multicolumn{5}{c||}{Easy Levels}  & \multicolumn{4}{c|}{Hard Levels} \\
			\hline
			& 0    & 1    & 2    & 3    & 4    & 5       & 6  & 7      & 8       \\
			\hline
			DQP     & -    & 0.83 & 1.06 & 0.95 & -    & \textbf{81.16}   & -  & \textbf{40.84}  & \textbf{7.07}    \\
			Random  & 0.36 & 0.36 & 0.19 & 0.37 & 0.49 & -       & -  & -      & -       \\
			Optimal & 0.16 & 0.15 & 0.15 & 0.12 & 0.14 & \textbf{-}       & -  & \textbf{-}      & \textbf{-}       \\
			BFS     & 0.03 & 0.04 & 0.05 & 0.04 & 0.04 & \textbf{-}       & -  & \textbf{-}      & \textbf{1789.6}  \\
			EHC     & 0.05 & 0.05 & 0.04 & 0.05 & 0.04 & 3.39    & -  & \textbf{-}      & \textbf{-}  \\
			\hline    
		\end{tabular}
	}
	
	\caption{Results obtained by each approach in Catapults. The symbol ``-" in the \emph{Optimal}, \emph{BFS} and \emph{EHC} rows represents a timeout (FF could not find a plan in 1 hour). In the \emph{DQP} and \emph{Random} rows it represents the corresponding approach was not able to solve that level (has a success rate of 0\%).}
\end{table}

\textbf{Results for BoulderDash}. Table 2 shows the results obtained by the different approaches in BoulderDash. The DQP model obtains plans which are approximately 23 \% longer than those obtained by the FF planner with the soft optimization options (\emph{BFS} and \emph{EHC} rows on top subtable of Table 2). The results obtained show this domain (game) poses difficulties for FF, which is only able to find the optimal plans for levels 1, 2 and 4, spending almost 7 minutes to do so for level 4. The BFS and EHC search strategies also present problems in this domain, particularly in hard levels. FF is only able to find a plan for level 0 using the BFS strategy (spending almost 4 minutes) and also spends more than 11 minutes to obtain a plan for level 3 with this same strategy. This shows FF has trouble solving even the easy levels. When we tried to solve the hard levels using FF, we could only find plans for levels 5 (with EHC) and 8 (with BFS), needing in both cases more than one minute of planning time. On the other hand, it can be observed that the DQP model can solve every level spending less than 2 seconds of total time, which accounts for both planning time and goal selection time. What is even more surprising is that the DQP model does not seem to spend more time in the hard levels than in the easy ones. If we take a look at the \emph{Random} row, we can observe that this model spends less time per level than the DQP model. This means that most of the time spent by the DQP model actually corresponds to the goal selection phase, i.e., every time the Goal Selection Module predicts the Q-value for a given \emph{(state, subgoal)} pair using the CNN. If we take this into consideration alongside with the fact that we are measuring \emph{total} time, which means that this time is actually split between every time the DQP model selects a subgoal, our approach drastically reduces the load of the planner for this domain, to an extent where FF can only solve less than half of the levels in reasonable time. At the same time, our approach obtains plans which are only slightly worse than those obtained by FF (using BFS or EHC), with only 23 \% more actions on average.

\textbf{Results for IceAndFire}.If we now take at look at table 3, we can observe that FF solves this domain a lot better than BoulderDash, being able to find the optimum plan for every level (although it spends around 10 seconds in levels 6, 7 and 8). Both the BFS and EHC methods solve all the easy levels almost instantly. Regarding the hard levels, EHC is able to solve them easily too and so does BFS, although it needs more than 5 seconds to solve levels 6, 7 and 8. As with BoulderDash, the DQP model spends around 1 second per level, regardless of its complexity. If we now focus on the quality (number of actions) of the plans obtained, it can be observed that the DQP model obtains plans which are, on average, as good as the ones obtained by EHC (only 2 \% longer on average) and only slightly worse than those obtained using BFS (17 \% longer on average). This shows our approach performs even better in this domain than in BoulderDash although all the levels are simple enough to be solved by FF quickly (except for levels 6, 7 and 8, for which BFS spends some more time).

\textbf{Results for Catapults.} Table 4 shows the results obtained for Catapults. This game is the hardest of the three by far, since for each level the subgoals (catapults) must be pursued in a very specific order or otherwise the agent will die. If we take a look at the success rate of the DQP model, we can appreciate it has trouble solving this game. On average, the DQP model obtains a sucess rate of 16 \% per level, which means it is able to solve 16 \% of the levels on average. This might seem low, but the Random model obtains a success rate of 7 \% per level, so the success rate of the DQP model is actually more than twice higher than the one obtained by Random model. This shows how hard this domain really is. If we now observe the results obtained by FF, we can see it is able to solve the easy levels without complications. However, when it comes to the hard levels, only EHC is able to solve level 5. Levels 6 and 7 can't be solved by FF (in one hour's time) with any search strategy, and level 8 can only be solved using BFS, spending almost half an hour. As with the other two domains, it seems DQP can solve the hard levels (except for level 6 for which it obtains a success rate of 0 \%) although it spends 43 seconds on average. This happens because these levels contains a lot of catapults (subgoals)  and that, besides the fact that DQP makes a lot of errors while selecting subgoals, means that the planner is called many more times than for the rest of the levels. If we now take a look at the length of the plans obtained, we can see that plans obtained by DQP are on average as good as those obtained by FF on the easy levels. For the hard levels, however, the plans obtained by DQP are longer than those obtained with FF, although level 7 can only be solved by DQP.

In the light of the results obtained, we can state that our approach obtains plans in the BoulderDash and IceAndFire domains of almost the same quality (length) as those obtained using classical planning, i.e., the FF planner. We have proved that, as the complexity of the problems to solve increases, the DQP model is able to solve them spending much less time than FF, to a point where for really complex problems FF fails to provide a solution in reasonable time (even with no optimization options involved). In Catapults, our approach fails to solve the levels most of the time. We have seen this is because this domain is really complex, as the success rate obtained by the Random Model shows. Due to this, even though the DQP model is able to obtain much better results than the Random Model, this is not enough for solving this domain reliably. For this reason, this domain must be solved using FF for simple levels although, as mentioned before, when the complexity of the levels increase FF is not able to solve this domain either. The results obtained by the DQP model in these domains seem to show that our approach is able to obtain good results, i.e., plans of good quality while spending little planning time, in domains of different kind, with the exception of domains where subgoals must be achieved in a very strict order, i.e., only a few of the different subgoal permutations correspond to a valid way of solving the level. However, even for these domains, it should be possible to obtain acceptable results by training the model on a bigger dataset.

\section{Related Work}
The use of Neural Networks (NN) in Automated Planning has been a topic of great interest in recent years. Some works have applied Deep Q-Learning to solve planning and scheduling problems as a substitute for online search algorithms. \cite{shen2017deep} uses Deep Q-Learning to solve the \emph{ship stowage planning problem}, i.e., in which slot to place a set of containers so that the slot scheme satisfies a series of constraints and optimizes several objective functions at the same time. \cite{mukadam2017tactical} also employs Deep Q-Learning, but this time to solve the \emph{lane changing problem}. In this problem, autonomous vehicles must automatically change lanes in order to avoid the traffic and get to the exit as quickly as possible. Here, Deep Q-Learning is only used to learn the long-term strategy, while relying on a low-level module to change between adjacent lanes without collisions. In our work, we also employ Deep Q-Learning but, instead of using it as a substitute for classical planning, we integrate it along with planning into our planning and acting architecture. Also, we do not focus on solving a specific problem but rather create an architecture which we hypothesize it is generalizable across a wide range of game domains.

There are other works which use neural networks to solve planning problems but, instead of relying on RL techniques such as Deep Q-Learning, train a NN so that it learns to perform an \emph{explicit planning process}. \cite{toyer2018action} proposes a novel NN architecture known as \emph{Action Schema Networks} (ASNet) which, as they explain in their work, \emph{are specialised to the structure of planning problems much as Convolutional Neural Networks (CNN) are specialised to the structure of images}. \cite{tamar2016value} uses a CNN that performs the computations of the value-iteration (VI) planning algorithm \cite{bellman1957dynamic,bertsekas2015dynamic}, thus making the planning process differentiable. This way, both works use NN architectures which \emph{learn to plan}. 

These NNs are trained on a set of training problems and evaluated on different problems of the same planning domain, showing better generalization abilities than most RL algorithms. \cite{tamar2016value} argues that this happens because, in order to generalize well, NNs need to learn an \emph{explicit planning process},  which most RL techniques do not. Although our architecture does not learn to plan it does incorporate an off-the-shelf planner which performs explicit planning. We believe this is why our architecture shows good generalization abilities.

Neural networks have also been applied to other aspects of planning. For instance, \cite{dittadi2018learning} trains a NN that learns a planning domain just from visual observations, assuming that actions have \emph{local} preconditions and effects. The learnt domain is generalizable across different problems of the same domain and, thus, can be used by a planner to solve these problems.

There exist several techniques which facilitate the application of Automated Planning in real-time scenarios, such as Goal Reasoning \cite{aha2015goal}, Anytime Planning \cite{richter2010lama}, Hierarchical Planning (e.g., HTN \cite{georgievski2015htn}) and domain-specific heuristics learned using ML \cite{yoon2008learning}. \cite{guzman2012pelea} presents PELEA, a domain-independent, online execution architecture which performs planning at two different levels, \emph{high} and \emph{low}, and is able to learn domain models, low-level policies and planning heuristics. \cite{mcgann2008deliberative} proposes T-REX, an online execution system used to control autonomous underwater vehicles. This system partitions deliberation across a set of concurrent \emph{reactors}. Each reactor solves a different part of the planning problem and cooperates with the others, interchanging goals and state observations.

In this work, we have proposed an architecture which uses Goal Reasoning as the method for interleaving planning and acting. \cite{jaidee2012learning} proposes a Goal Reasoning architecture which uses Case-Based Reasoning \cite{kolodner2014case} and Q-Learning in order to learn to detect discrepancies, associate discrepancies to new goals and learn policies that achieve the selected goals. In our work, we have focused on learning to select subgoals, using a NN (integrated into the Deep Q-Learning algorithm) instead of traditional Q-Learning in order to give our architecture the ability to generalize. For this reason, we believe our approach scales better when applied to big state spaces than the one proposed in \cite{jaidee2012learning}. In future work, we plan to extend our architecture so that it is also able to learn new subgoals.

\cite{bonanno2016selecting} employs an architecture that does use a NN, concretely a CNN, to select subgoals for navigating a maze in the game known as \emph{Minecraft}. When a subgoal must be selected, the CNN receives an image of the current state of the game, which is used to decide the most suitable subgoal for that state. Unlike our work, a hard-coded expert procedure is used to teach the CNN which subgoal must be selected in each state. As Bonanno et al. recognise, this approach transforms the problem into a classification task, instead of a RL one. Furthermore, the set of eligible subgoals are always the same four regardless of the state of the game. In our work, the compound subgoal $G$ is different for each game state and can contain a different number of single subgoals $g \in G$ to choose from.

Finally, it is worth to mention previous disruptive work on Deep RL \cite{mnih2015human} that addresses how to learn  models to control the behavior of reactive agents in ATARI games. As opposite to this work, we are interested in addressing how deliberative behaviour (as planning is) can be improved by mainstream techniques in Machine Learning. This is one of the main reasons we chose the GVGAI video game framework, since it provides an important repertory of video games where deliberative behaviour is mandatory to achieve a high-level performance.

\section{Conclusions and Future Work}
We have proposed a goal selection method which learns to select subgoals with Deep Q-Learning in order to interleave planning and acting. We have tested our architecture on three different GVGAI games, using different levels for training and testing. We have compared our approach with a classical planner, measuring both the quality (length) of the plans and the time spent to obtain them.

We have proved our approach is able to obtain plans of similar quality to those obtained by a classical planner, needing on average much less time to solve complex problems (levels). We have also shown our DQP model is applicable to domains (games) of different kind and presents good generalization properties when applied to new levels. Unlike our model, most RL techniques can't generalize well \cite{zhang2018study}. At the same time, the original DQN paper \cite{mnih2013playing} utilizes a training dataset of 10 million samples, whereas we only use around 50000 samples to train our model.

We believe the reason behind all of this is that, with our approach, we are actually splitting the planning problem into two parts. On the one hand, we use RL (Deep Q-Learning specifically) to select subgoals, which can be interpreted as a form of high-level planning. On the other hand, we use a classical planner (FF) to achieve each selected subgoal, which can be viewed as a form of low-level planning. This way, the complexity of the problem to solve is split and shared between the RL algorithm and the planner. So, the same as the load of the planner is greatly reduced (which manifests as much smaller planning times), the Deep Q-Learning algorithm also obtains way better results (better generalization while being more sample-efficient) than it would normally do without the planner's help. We believe this synergy is the key element of our approach.

One limitation of our work is that, in order to apply our architecture to a new game, we need to manually create its associated domain. In future work, we intend to make use of the method detailed in \cite{ignacio} to automatically obtain PDDL domains from VGDL game descriptions. We also plan to learn to formulate goals, in order to achieve truly generalization across domains. Lastly, we plan to augment our approach so that it can be used in non-deterministic environments. We believe this should be as easy as training our DQP model to predict the uncertainty or risk associated with a subgoal, in addition to the length of the corresponding plan.

\section{Acknowledgments}
Financial support tbd.


\bibliography{References}
\bibliographystyle{aaai}
\end{document}